\documentclass[conference]{IEEEtran}
\IEEEoverridecommandlockouts
\usepackage{cite}
\usepackage{amsmath,amssymb,amsfonts}
\usepackage{algorithm}
\usepackage{algpseudocode}
\usepackage{graphicx}
\usepackage{textcomp}
\usepackage{xcolor}
\usepackage{enumitem}
\usepackage{subcaption}
\usepackage{booktabs}
\def\BibTeX{{\rm B\kern-.05em{\sc i\kern-.025em b}\kern-.08em
    T\kern-.1667em\lower.7ex\hbox{E}\kern-.125emX}}
\begin{document}

\algnewcommand\algorithmicpsdo{\textbf{Central server do:}}
\algnewcommand\PSDO{\item[\algorithmicpsdo]}
\algnewcommand\algorithmicclientdo{\textbf{Clients do:}}
\algnewcommand\ClientDO{\item[\algorithmicclientdo]}
\algnewcommand\algorithmicblank{\textbf{}}
\algnewcommand\Blank{\item[\algorithmicblank]}
\algblockdefx{ForAllP}{EndForAllP}[1]%
  {\textbf{for all }#1}%
  {\textbf{end for}}

\algnewcommand\algorithmictrainlocally{\textbf{TrainLocally($k$, $w_0$):}}
\algnewcommand\TrainLocally{\item[\algorithmictrainlocally]}

\title{FedSyn: Synthetic Data Generation using Federated Learning\\
}


\author{Monik Raj Behera\textsuperscript{1}, Sudhir Upadhyay\textsuperscript{1}, Suresh Shetty\textsuperscript{1}, Sudha Priyadarshini\textsuperscript{1},  Palka Patel\textsuperscript{1}, Ker Farn  Lee\textsuperscript{1}\\

    \\\small{\texttt{\{monik.r.behera,sudhir.x.upadhyay,suresh.shetty,sudha.priyadarshini\}}}\\
    \small{\texttt{@jpmorgan.com}}
      \\\\{\textsuperscript{1}Onyx by J.P. Morgan}}

\maketitle

\begin{abstract}
As Deep Learning algorithms continue to evolve and become more sophisticated, they require massive datasets for model training and efficacy of models. Some of those data requirements can be met with the help of existing datasets within the organizations.
Current Machine Learning practices can be leveraged to generate synthetic data from an existing dataset.  Further, it is well established that diversity in generated synthetic data relies on (and is perhaps limited by) statistical properties of available dataset within a single organization or entity. 
The more diverse an existing dataset is, the more expressive and generic synthetic data can be. However, given the scarcity of underlying data, it is challenging to collate big data in one organization.  The diverse, non-overlapping dataset across distinct organizations provides an opportunity for them to contribute their limited distinct data to a larger pool that can be leveraged  to further synthesize. Unfortunately, this raises data privacy concerns that some institutions may not be comfortable with.

This paper proposes a novel approach to generate synthetic data - \textit{FedSyn}. FedSyn is a collaborative, privacy preserving approach to generate synthetic data among multiple participants in a federated and collaborative  network. 
FedSyn creates a synthetic data generation model, which can generate synthetic data consisting of statistical distribution of almost all the participants in the network. 
FedSyn does not require access to the data of an individual participant, hence protecting the privacy of participant's data. 
The proposed technique in this paper leverages federated machine learning and generative adversarial network (GAN) as neural network architecture for synthetic data generation. 
The proposed method can be extended to many machine learning problem classes in finance, health, governance, technology and many more.
\end{abstract}


\section{Introduction}
Back in 2017, The Economist published a story titled, "The world's most valuable resource is no longer oil, but data."  Since then data has continued to be one of the most critical and valuable asset in industry today, where we see innovative and complex machine learning algorithms and architectures, which are dependent on a plethora of data for training, testing, mining and designing `data-centric' statistical algorithms\cite{alvarez2021towards}. 
In many organizations, big data is mined with robust data engineering pipelines, which allows them to create sufficient datasets to train machine learning and deep learning models.
Scarcity of data can be a big challenge to application of machine learning for organizations or entities which are not able to mine big data. 
Though transfer learning\cite{weiss2016survey} may help, it does not cater to all the needs as they are generally designed over public data. 
In few cases, even organizations with abundant data may find some bias in their dataset because of various reasons, resulting in biased and non-generic machine learning models with limited variance in available data. 

GAN\cite{goodfellow2014generative} have emerged as one of the prominent neural network architectures to generate synthetic data. The architecture can be modified to accommodate various data signals. Since discriminator and generator models compete with each other in an adversarial game to generate synthetic data as close as possible to original data, the quality of synthetic data is directly dependent on training data used. Generative Adversarial Networks trained on available data can help generate data required for training machine learning models in organizations. 

In financial, health care and Internet of Things industries, recent trends\cite{9409764, rieke2020future, 9153560, kumar2021federated, long2020federated, zheng2021applications} have indicated an increase in collaboration among organizations to benefit from data of their peers in their respective industries. To draw a hypothetical scenario, different financial organizations have varying concentration of their business operations in different geographical distributions. This in turn results in the concentration of a specific dataset in those locations. Localized regional data would certainly be valuable to other regions that can leverage it for training their local models for developing intelligent solutions, including anomaly detection, potential payment fraudulent in case of cross border payments etc. However, considering data privacy laws and governance on data accessibility within industry it may be challenging to share such data across regions.  In such cases, it may be possible to share the local models across regions that can augment other regional models. To extend it further, the proposed method in this paper can be implemented for greater research opportunities to obtain powerful and global machine learning models.

\subsection{Novel Contribution in Current Work}
Industries, research groups and individuals would benefit from machine learning models using collaboration, which will bridge the gap of \textit{data scarcity} and \textit{data bias}\cite{kaur2019systematic}. Though collaboration is required, \textit{data privacy} must be honoured. 
Generating synthetic data is a strategic way to proceed, where challenges to data scarcity could be solved, but data bias would still be observed, since synthetic data generation models would be trained on existing data only. The experiments in this paper demonstrate generation of synthetic data in a privacy preserving and collaborative manner.
\begin{enumerate}[label=\alph*.]
    \item Data Scarcity
    \item Data Bias
    \item Data Privacy
\end{enumerate}

In order to try and solve the above-mentioned challenges, \textit{FedSyn} framework has been proposed in this paper. FedSyn proposes a novel method, in which federated learning, synthetic data generation using GAN and differential privacy are combined as a framework. For solving the challenge of data scarcity, generative adversarial networks are being used to generate synthetic data after being trained upon existing data. Federated learning\cite{konevcny2016federated} provides privacy of data as well as that of underlying data of participants in collaborative network, with aggregation algorithm executed on \textit{non-IID data}\cite{zhu2021federated}. This essentially tries to solve the problem of data bias, as aggregated models from federated server builds on learning from individual participants. To extend data privacy and anonymity guarantee even further, FedSyn implements differential privacy by adding Laplacian noise\cite{sarathy2011evaluating} to model parameters.

\section{Related Work}
Synthetic data generation has been an active area of research and exploration, both for research groups and industries. Synthetic data is not only used for training machine learning, but software engineering industry also depends on synthetic data for testing, and other business use cases. With the introduction of GAN, it has become de facto method to generate synthetic data signals like image, uni-variate data, multivariate data, audio, video, etc.

\subsection{Generative Adversarial Network}
As discussed in \cite{emami2018generating}, GANs have been quite popular in the domain of synthetic data generation. They are being actively researched in areas of data and image augmentation, generating synthetic images for medical domain, data for extensive training and many more.
In \cite{creswell2018generative}, authors have discussed the common design patterns, use cases and various architectures in GANs - convolutional GAN, conditional GAN, tabular GAN, adversarial auto-encoders. 

\subsection{Federated Learning}
Horizontal federated learning\cite{yang2019federated} with non-IID data is the key consideration in this paper, where participants have varied data distribution and properties. A majority of the enterprise networks for federated learning are constrained. This ensures homogeneous model parameters and architecture for all the participating clients in federated learning network. In real world scenarios, where data originate from different organizational entities, covariate shift, prior probability shift, concept shift and unbalanced data size are common technical challenges\cite{li2020federated}.

\subsection{Differential Privacy}
In \cite{geng2015optimal}, authors have discussed through numerous experiments, that Laplacian noise performs better over Gaussian noise by significantly reducing noise amplitude and noise power in all privacy regimes. This also provides an effective noise distribution, which does not reduce the performance of aggregated machine learning model as compared to Gaussian noise, since Gaussian distribution has higher standard deviation (and noise spread).
In \cite{xu2019ganobfuscator}, authors have discussed methods of adding noise to gradients of the neural network while training. This is to obfuscate generator, which provides strict differential privacy guarantee. 
In \cite{jordon2018pate}, Private Aggregation of Teacher Ensemble framework is implemented with GAN, which further guarantees tight differential privacy by diluting the effect of any individual sample.

\subsection{Federated Learning and Synthetic Data}
In \cite{9054559}, authors have showed usage of federated learning for GANs, with Gaussian noise in data during individual participant training (local training), which would protect the privacy of participants. This work has also provided theoretical guarantee of $(\epsilon,\delta)$-differential privacy. In \cite{goetz2020federated}, authors have shown the usage of synthetic data to enhance the communication rounds in federated learning. The participants in federated learning are sending synthetic data, generated from original data, which will be used for training on federated server. This ensures data privacy and optimizes communication overhead, which can be really effective in complex real-world scenarios. 

\section{Federated Learning-based Synthetic Data Generation}
For a given problem specification, every individual participant will be executing GAN based synthetic data generator. This will generate synthetic data, which must be deployed for usage, monitored for various possible challenges like data drift, temporal drift, distribution shift, etc. Since the local GAN process is independent of federated learning network, it should follow the general machine learning pipeline practiced across industry, which is comprised of data engineering, model monitoring and feedback for supervised models\cite{vartak2016modeldb}. 

\begin{figure*}[ht]
\vskip 0.2in
\begin{center}
\centerline{\includegraphics[width=0.65\linewidth]{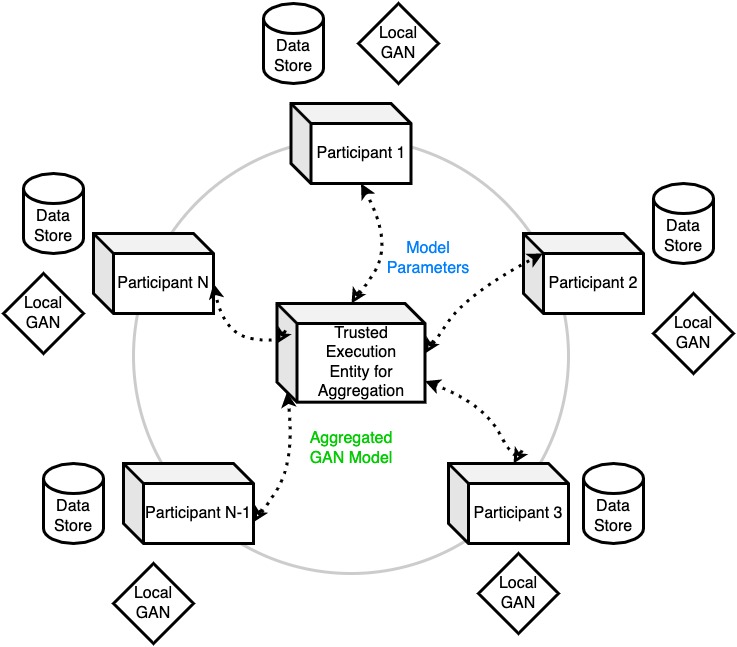}}
\caption{In the above architecture diagram, individual nodes are running data engineering jobs with processes to serve and re-train data models. The participants are connected over a secure network, with many-to-one star topology to share model parameters. Each individual participant consists of data storage and local GAN process}
\label{overview}
\end{center}
\vskip -0.2in
\end{figure*}

\begin{figure}[ht]
\vskip 0.2in
\begin{center}
\centerline{\includegraphics[width=\columnwidth]{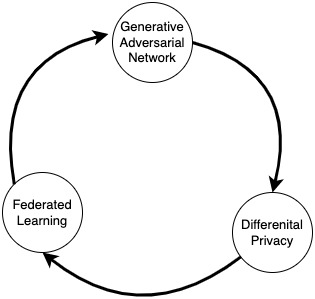}}
\caption{In the above diagram, three core components are depicted - General adversarial network deep neural network at local participant level, differential privacy added at local participant level and aggregation layer, and finally federated learning over the complete network}
\label{core_components}
\end{center}
\vskip -0.2in
\end{figure}

\textit{FedSyn} is a privacy preserving method of generating synthetic data in a collaborative manner. The method is suitable for enterprise networks, where individual participants have ample resources to train complex neural networks. Network participant with sufficient data for deep neural networks have a better chances of contributing to federated aggregate model, as compared to participants with resource constraints. The framework is dependent on creating a collaborative and trusted network of servers, communicating model parameters in every federated learning communication round with trusted execution entity over secure channels, as depicted in Figure \ref{overview}. FedSyn is comprised of three core components in one round, as depicted in Figure \ref{core_components}.

\subsection{Generative Adversarial Network}
In the current work, GAN is used to generate synthetic data at individual participant level. As with every GAN, it consists of two separate neural network models - 
\begin{enumerate}
    \item Generator
    \item Discriminator
\end{enumerate}
In case of GAN training, the data is prepared and fed into discriminator network for training. Further, random latent space is generated with Gaussian distribution, which is then trained further as an adversarial game with discriminator network to achieve an objective, where generator generates latent space which are classified as real by discriminator network. The epochs for this training can be decided empirically, as optimal number of epochs are dependent on neural network architecture and data used. 

\begin{figure*}[ht]
     \centering
     \begin{subfigure}[b]{\textwidth}
         \centering
         \includegraphics[width=\textwidth]{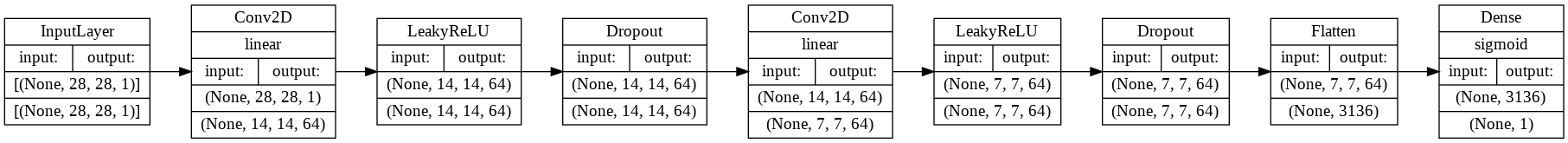}
         \caption{In above figure, the Discriminator neural network architecture is depicted in Tensorflow Keras format, specific to this paper. The top layer is input layer, followed by 2D Convolution layer and LeakyRelu layer followed by Dropout layer. The subsequent layer is 2D Convolution layer with LeakyRelu with reduced kernel size, followed by Dropout layer. The next layer is Flatten layer. The final layer is Dense layer with Sigmoid layer}
     \end{subfigure}
     \begin{subfigure}[b]{\textwidth}
         \centering
         \includegraphics[width=\textwidth]{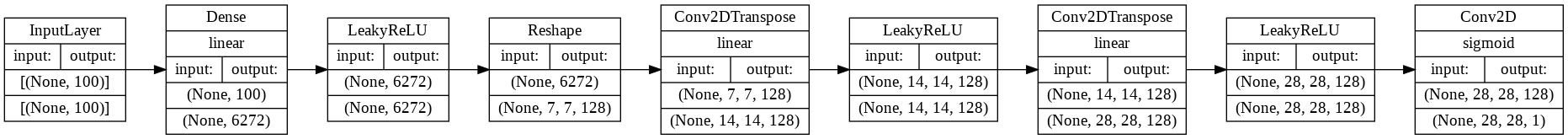}
         \caption{In above figure, the Generator neural network architecture is depicted in Tensorflow Keras format, specific to this paper. The top layer is input layer, followed by Dense layer and LeakyRelu layer followed by Reshape layer. The subsequent layer is 2D Convolution transpose layer with LeakyRelu, again followed by 2D Convolution transpose layer with LeakyRelu with doubled kernel size. The final layer is 2D Convolution layer in desired shape of expected data with Sigmoid layer}
     \end{subfigure}
        \caption{GAN model's generator and discriminator's neural network architecture}
        \label{gan_network_plot}
\end{figure*}

The discriminator network is essentially a binary classifier to classify between real and fake images. Figure \ref{gan_network_plot}.a depicts the discriminator network which is used for the paper. Discriminator can be modified to obtain various other architectures depending on the objective and data. 

Generator network is a feature extractor, which generates latent space following specific distribution learnt during training. Similar to discriminator network, generator network can follow various architectures depending on the use case. Figure \ref{gan_network_plot}.b shows the generator network used in the current work. 

\subsection{Differential Privacy}
In \cite{dwork2008differential}, differential privacy is defined as a method to protect privacy of underlying data and statistical properties, which can be potentially leaked from models. The core idea is to add noise onto training data, which would be added to model parameters and learnt indirectly, since the data used for training is modified. 

In the current work, differential privacy is implemented by adding noise to model parameters, rather than data. Doing this adds noise to generator weight layers, which are responsible for generating latent space for synthetic images\cite{odena2018generator}. Adding noise to model parameters of generator adds noise to latent space in multi-dimensional hyper-sphere, effecting synthetic images itself, hence providing similar benefits as of adding noise to data itself. This approach also makes it computationally favourable for all the participants. Laplacian noise is added to the model parameters before they are sent from participant server to aggregation server. Use of Laplacian noise is favourable over Gaussian noise, as the resultant model parameters are quantitatively less distorted compared to Gaussian noise. Even after that, privacy guarantees are better for Laplacian noise\cite{sarathy2011evaluating}. Below equation shows how the noise from each participant is accumulated at federated aggregated server. One can observe that noise from each participant is added together and propagated further in aggregated model parameters, and subsequent federated learning rounds. 

\begin{equation} \label{eq_wavg}
    W_{G} = \sum_{i}^{N}\frac{w_{i}p_{i}}{\sum_{i}^{N}p_{i}} + \sum_{i}^{N}\frac{\mathcal{L}_{i}p_{i}}{\sum_{i}^{N}p_{i}}
\end{equation}

where $W_{G}$ is aggregated model parameter, $N$ is total number of participants, $w_{i}$ is model parameter of $i^{th}$ participant, $p_{i}$ is scaling weight of $i^{th}$ participant, $\mathcal{L}_{i}p_{i}$ is Laplacian noise for $i^{th}$ participant. 

To extend further, Laplacian noise is also added during aggregation for increased privacy. This enhances Equation \ref{eq_wavg} as below

\begin{equation} \label{eq_wavggl}
    W'_{G} = \sum_{i}^{N}\frac{w_{i}p_{i}}{\sum_{i}^{N}p_{i}} + \sum_{i}^{N}\frac{\mathcal{L}_{i}p_{i}}{\sum_{i}^{N}p_{i}} + \mathcal{L}_{G'}
\end{equation}

where $\mathcal{L}_{G'}$ is Laplacian noise added during aggregation of participant's model parameters and $W'_{G}$ is the enhanced model parameter after aggregation, with additive noise accumulation.

In \cite{xu2019ganobfuscator, jordon2018pate}, the authors have discussed and shown privacy guarantees of using differential privacy on various layers of neural network architecture and training. Though data leakage is a core challenge in synthetic data generation, which requires continued research, our work, inspired by \cite{xu2019ganobfuscator, jordon2018pate} does introduce noise to model parameters, which may mitigate data leakage to a certain extent. 

\subsection{Federated Learning}
As described in \cite{bonawitz2019towards}, the current federated learning framework consists of a trusted and secure aggregator server. The federated learning rounds are orchestrated by aggregation server through time bound triggers. In every round, participants share their model parameters to the aggregator server through secure and anonymous channels. Due to the nature of communication, it follows star topology\cite{dowd1991random} for network communication. Aggregator server responsible for aggregation of all the local participant's model parameters using \textit{FedAvg} algorithm \cite{li2019convergence}. 

A key enhancement done in FedSyn architecture for averaging algorithm is to add Laplacian noise during FedAvg aggregation. The enhanced algorithm is depicted in Algorithm \ref{fedavg_fedsyn}. Laplacian noise is added after the weighted averaging of model parameters is completed. This further extends the required privacy of the model, which will protect against adversarial attacks and prevent participant identity leakage. 

\begin{algorithm}[ht] 
	\caption{\texttt{FedSyn - FederatedAveraging}. In the cluster there are $N$ clients in total, each with a learning rate of $\eta$. The set containing all clients is denoted as $S$, the communication interval is denoted as $E$, and the weights of clients is denoted as $P$. The Laplacian noise is denoted as $\mathcal{L}_{noise}(\mu, \lambda)$, where $\mu$ denotes mean of Laplacian distribution and $\lambda$ denotes exponential decay parameter of Laplacian distribution}
	\label{fedavg_fedsyn} 
	\begin{algorithmic} 
	    \PSDO
	        \State Initialization: global model $w_0$.
	        \For {each global iteration $t \in {1, ..., iteration}$}
	            \ForAllP {each client $k \in S$}
	                \State \# Get clients improved model.
	                \State $w_{t+1}^{k} \leftarrow TrainLocally(k, w_t)$
	            \EndForAllP
	            \State \# Update the global model.
	            \State $w_{t+1} \leftarrow \mathcal{L}_{noise}(\mu, \lambda) + \sum_{k=0}^{N} p_{k}w_{t+1}^{k}$
	        \EndFor
    \Blank
	\TrainLocally
	    \For {each client iteration $e \in {1, ..., E}$}
	        \State \# Do local model training.
	        \State $w_{e} \leftarrow w_{e-1} - \eta \nabla F(w_{e-1})$
	    \EndFor
	    \State \Return $w_{E}$
	\end{algorithmic} 
\end{algorithm}

\section{Experiments}
In the current work, the performance of synthetic data generated with \textit{FedSyn} method has been captured on both MNIST\cite{deng2012mnist} and CIFAR10\cite{krizhevsky2014cifar} public dataset. Both the datasets are widely used to perform federated learning based experiments, since it contains well sampled data elements for 10 labels. For the current work, MNIST and CIFAR10 data are divided into 3 parts based on labels, in separate experiments. In each experiment, part 1 denotes all the images from dataset which have labels 0, 1 and 2 with train size of 15000. Part 2 denotes images for labels 3, 4, 5, and 6 with train size of 20000. Part 3 denotes images for labels 7, 8 and 9 with train size of 15000. This sampling allows to have 3 different simulated participants in federated learning network, in non-IID manner. Given the sample size, the importance weight factor $p_{k}$ mentioned in Algorithm \ref{fedavg_fedsyn} is 0.3, 0.4 and 0.3 respectively. 

For local model training of GANs, 4 core, 16 GB memory based \textit{AWS ec2} machines were used for each simulated participant and aggregator server. Each image from dataset has resolution of 28x28. For the current work, architecture of used generator neural network is described in Figure \ref{gan_network_plot}.b and architecture of discriminator neural network is described in Figure \ref{gan_network_plot}.a. Neural networks are implemented using Tensorflow framework\cite{tensorflow2015-whitepaper}. For training, batches of 256 have been used with loss function as binary cross entropy\cite{ho2019real}. Adam optimizer\cite{kingma2017adam} with learning rate of 0.0002 and exponential decay rate of 0.5 is used.

\begin{figure}[ht]
\vskip 0.2in
\begin{center}
\centerline{\includegraphics[width=\columnwidth]{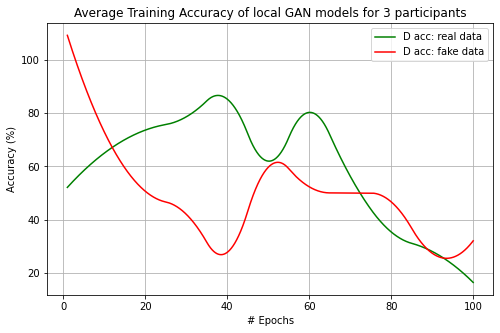}}
\caption{In above figure, average training accuracy (across both the datasets and all the simulated participants), in percentage, is plotted across number of epochs. Red line depicts accuracy for classifying fake data points and Green line depicts accuracy for classifying real data points.}
\label{avg_tr_acc}
\end{center}
\vskip -0.2in
\end{figure}

The training accuracy for each participants is taken and mean is computed for each epoch checkpoint (10 epochs per checkpoint). Training is done for 100 epochs. Figure \ref{avg_tr_acc} shows the average (average over all the simulated participants, in both the datasets' experiments) training accuracy of discriminator for classifying real and fake images accurately. From the plot, it is observed that higher number of epochs does not necessarily guarantee better accuracy of models. This phenomenon may occur because of higher learning rate and going beyond global minima of cost function\cite{zhang2018generalized}. In the experiment, the models after 50 epochs have been considered as final model for each participant, whose model parameters are sent to aggregator server. 

\begin{figure*}[ht]
     \centering
     \begin{subfigure}[b]{0.3\textwidth}
         \centering
         \includegraphics[width=\textwidth]{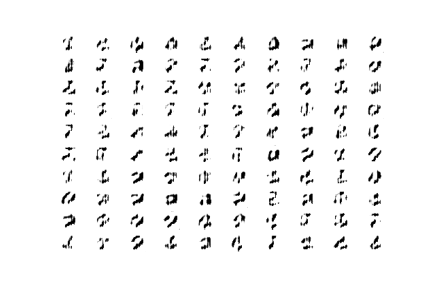}
         \caption{Generated images after 10 epochs}
     \end{subfigure}
     \begin{subfigure}[b]{0.3\textwidth}
         \centering
         \includegraphics[width=\textwidth]{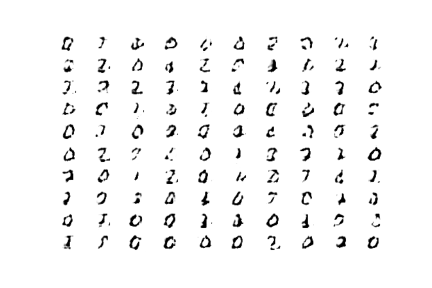}
         \caption{Generated images after 20 epochs}
     \end{subfigure}
     \begin{subfigure}[b]{0.3\textwidth}
         \centering
         \includegraphics[width=\textwidth]{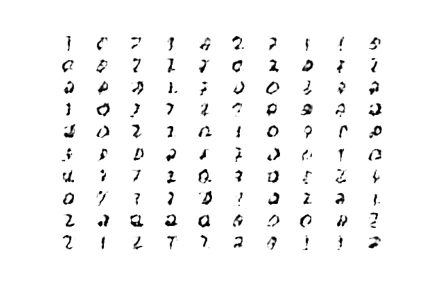}
         \caption{Generated images after 30 epochs}
     \end{subfigure}
     \begin{subfigure}[b]{0.3\textwidth}
         \centering
         \includegraphics[width=\textwidth]{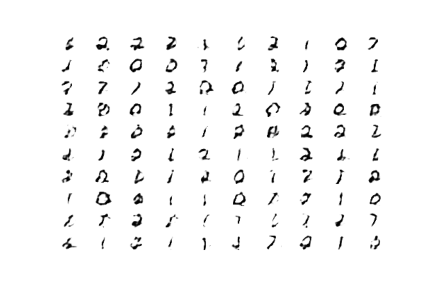}
         \caption{Generated images after 40 epochs}
     \end{subfigure}
     \begin{subfigure}[b]{0.3\textwidth}
         \centering
         \includegraphics[width=\textwidth]{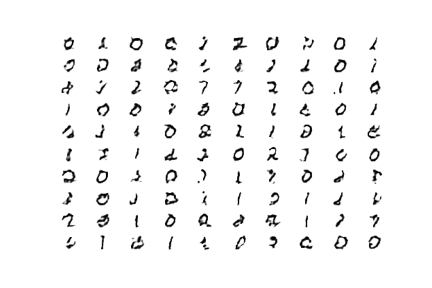}
         \caption{Generated images after 50 epochs}
     \end{subfigure}
        \caption{Qualitative analysis of generated images for MNIST data, over number of epochs during training of local GAN model for Participant 1}
        \label{figure_1_learning}
\end{figure*}

As discussed in \cite{Shmelkov_2018_ECCV}, measuring quantitative performance for GAN generated images is still an active research area, with no standard method. Currently, in order to measure performance of images generated with GAN, one can employ problem specific and data specific methods, or can proceed with qualitative methods, involving subjective analysis of generated images. In Figure \ref{figure_1_learning}, the quality of synthetically generated images for MNIST data can be qualitatively observed improving over number of epochs during training. 

\begin{figure*}[ht]
     \centering
     \begin{subfigure}[b]{0.2\textwidth}
         \centering
         \includegraphics[width=\textwidth]{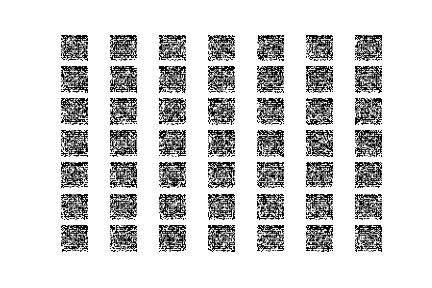}
         \caption{$\lambda=10^{0}$}
     \end{subfigure}
     \begin{subfigure}[b]{0.2\textwidth}
         \centering
         \includegraphics[width=\textwidth]{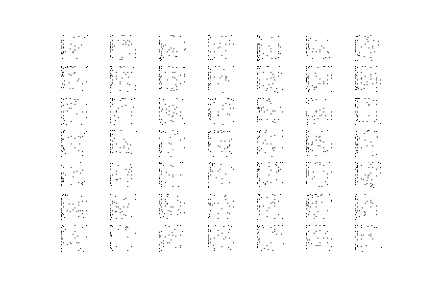}
         \caption{$\lambda=10^{-1}$}
     \end{subfigure}
     \begin{subfigure}[b]{0.2\textwidth}
         \centering
         \includegraphics[width=\textwidth]{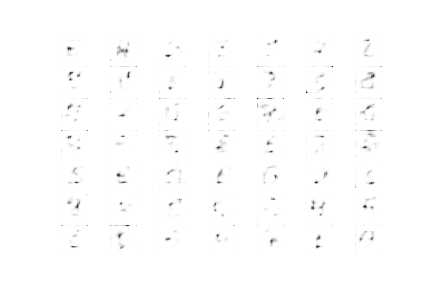}
         \caption{$\lambda=10^{-2}$}
     \end{subfigure}
     \begin{subfigure}[b]{0.2\textwidth}
         \centering
         \includegraphics[width=\textwidth]{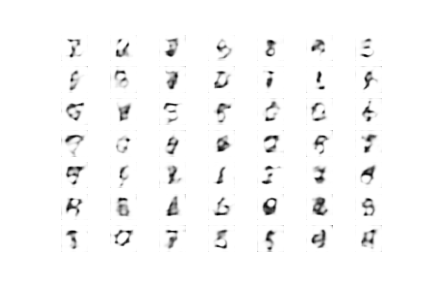}
         \caption{$\lambda=10^{-3}$}
     \end{subfigure}
     \begin{subfigure}[b]{0.2\textwidth}
         \centering
         \includegraphics[width=\textwidth]{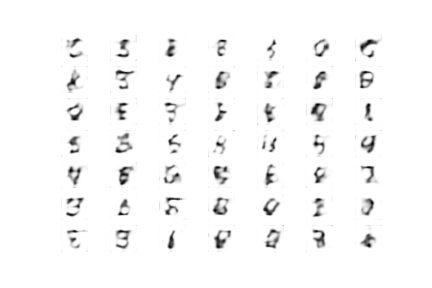}
         \caption{$\lambda=10^{-4}$}
     \end{subfigure}
     \begin{subfigure}[b]{0.2\textwidth}
         \centering
         \includegraphics[width=\textwidth]{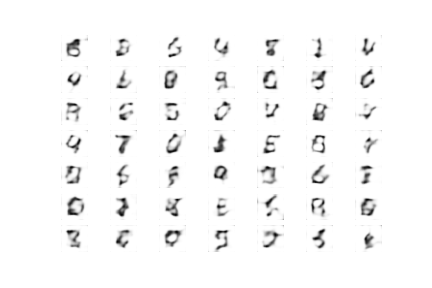}
         \caption{$\lambda=10^{-5}$}
     \end{subfigure}
     \begin{subfigure}[b]{0.2\textwidth}
         \centering
         \includegraphics[width=\textwidth]{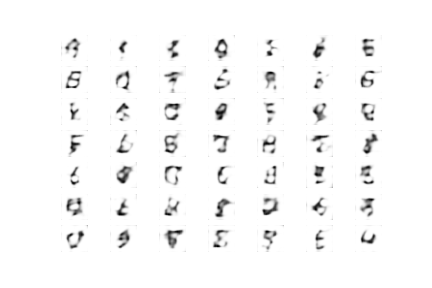}
         \caption{$\lambda=10^{-6}$}
     \end{subfigure}
        \caption{Qualitative analysis of synthetic images generated for MNIST data, from global model using aggregated model parameters over changing $\lambda$ of Laplacian noise. Laplacian noise is added to each local model parameters for differential privacy during federated learning.}
        \label{fig_lap_noise}
\end{figure*}

After the completion local GAN training, Laplacian noise\cite{sarathy2011evaluating} with $\mu=0$ and $\lambda$ as $10^{0}$,$10^{-1}$, $10^{-2}$, $10^{-3}$, $10^{-4}$, $10^{-5}$, $10^{-6}$ is added to model parameters.
Resultant model parameters from each participant are securely sent to aggregator server for aggregation, as described in Algorithm \ref{fedavg_fedsyn}. Figure \ref{fig_lap_noise} shows the effect of changing $\lambda$ of Laplacian noise added for differential privacy, over quality of images generated, for MNIST data. The synthetic images generated in Figure \ref{fig_lap_noise} are from global model, with aggregated model parameters from federated learning. It can be observed that quality of image does not change significantly after for $10^{-4}, 10^{-5}$ and $10^{-6}$. So higher value $\lambda$ can be chosen for the given problem, as higher the $\lambda$, higher is the degree of privacy maintained. 

\begin{figure}[ht]
\vskip 0.2in
\begin{center}
\centerline{\includegraphics[width=\columnwidth]{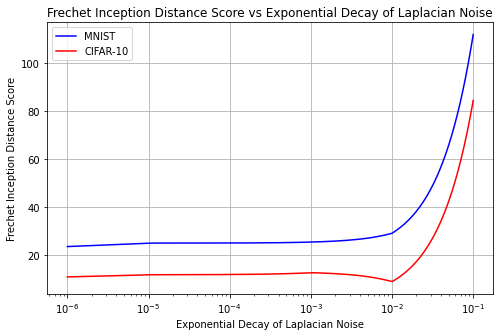}}
\caption{In above figure, the Frechet Inception Distance Score is plotted against exponential decay parameter of Laplacian noise added for differential privacy}
\label{fid_fig}
\end{center}
\vskip -0.2in
\end{figure}

Though quantitative metric for GANs is still an active area of research, Frechet Inception Distance (FID) \cite{Shmelkov_2018_ECCV} is a metric which calculates distance between feature vector of base images and synthetically generated images. FID scores have been also used several other works\cite{9054559, Shmelkov_2018_ECCV, hardy2019md} related to synthetic data generation using federated learning. Lesser the FID score, better is the quality of synthetic images generated. In our case, base images are the synthetic images generated by a GAN model, which is trained over all the MNIST and CIFAR10 data, centrally (in two separate experiments). Thus, central GAN model forms the upper bound for synthetic images generated with FedSyn. As it can be observed in Figure \ref{fid_fig}, with increasing value of $\lambda$ in Laplacian noise used for differential privacy, the value of FID score is also increasing. This essentially shows the deteriorating effect of generated images with increase in noise. FID score could be a great metric, which can help in deciding the optimal value of $\lambda$ for Laplacian noise. After observing Figure \ref{fig_lap_noise} and \ref{fid_fig}, it is evident that $\lambda = 10^{-4}$ is an optimal choice, where images generated are of acceptable quality and the degree of differential privacy is also sound. The results of hyper parameter $\lambda$ over both the datasets are also observed in Table \ref{table_fid}.

\begin{table}[t]
\caption{Frechet Inception Distance score of images generated from FedSyn, compared to images generated with centrally trained GAN model.}
\label{table_fid}
\vskip 0.15in
\begin{center}
\begin{small}
\begin{tabular}{|p{0.4\linewidth}|p{0.2\linewidth}|p{0.2\linewidth}|}
\toprule
$\lambda$ of Laplacian Noise & MNIST & CIFAR10 \\
\midrule
$10^{-6}$ & $23.43$ & $10.81$ \\
\midrule
$10^{-5}$ & $24.88$ & $11.71$ \\
\midrule
$10^{-4}$ & $24.91$ & $11.91$ \\
\midrule
$10^{-3}$ & $25.31$ & $12.58$ \\
\midrule
$10^{-2}$ & $28.96$ & $8.84$ \\
\midrule
$10^{-1}$ & $111.63$ & $84.21$ \\
\midrule
$10^{0}$ & $132.45$ & $97.74$ \\
\bottomrule
\end{tabular}
\end{small}
\end{center}
\vskip -0.1in
\end{table}

\begin{figure}[ht]
     \centering
     \begin{subfigure}[b]{0.4\textwidth}
         \centering
         \includegraphics[width=\textwidth]{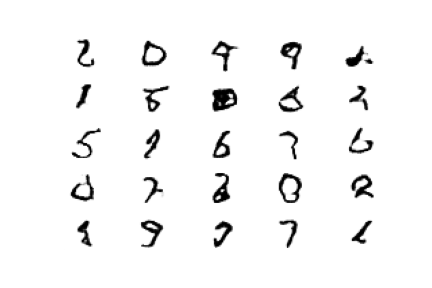}
         \caption{Synthetic Images generated from generator of GAN model, which is centrally trained over all the labels (without federated learning)}
     \end{subfigure}
     \begin{subfigure}[b]{0.4\textwidth}
         \centering
         \includegraphics[width=\textwidth]{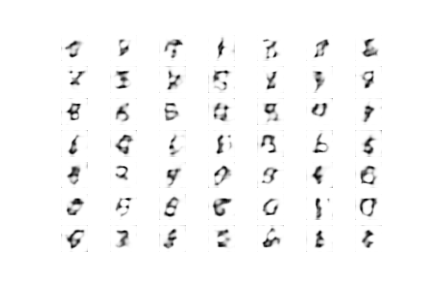}
         \caption{Synthetic Images generated from FedSyn generator model, converged model using federated learning}
     \end{subfigure}
        \caption{Qualitative analysis of quality of images generated from generator of GAN model, which is trained over all the labels versus synthetic images generated from generator of FedSyn model.}
        \label{fig_fedsyn_gan}
\end{figure}

To draw a comparison of quality of image generated through FedSyn, Figure \ref{fig_fedsyn_gan} shows synthetically generated images from GAN trained over all the 10 labels versus synthetic generated images from FedSyn model, for MNIST data. 
As it can be observed, the images generated from global generator model, created after using aggregated model parameters from federated learning round consists of images from all the 10 labels (handwritten digits). 
Whereas, individual participants have been trained on fraction of total labels. Also, aggregator server does not have access to all the data points, but the aggregated model parameters is capable of generating synthetic data from all the labels. 
This observation supports the claim of using FedSyn on non-IID data with Laplacian noise for differential privacy on local generator model parameters, which has been proposed in this paper. 

\section{Conclusion}
Growing need for data intensive, deep learning models across industries is the key driver for collaboration among enterprise and participants. The key challenges faced here are data privacy, data scarcity and data bias. 
In the current work, Privacy preserving method like federated learning along with synthetic data generation using GAN is implemented to create \textit{FedSyn} framework. 

FedSyn tries to solve the data privacy challenge with federated learning where the original data is never shared for GAN training, but model parameters are shared, after applying differential privacy with Laplacian noise.
Federated learning with non-IID data among participants of network, minimizes the risk of data bias while aggregating the local model parameters. 
Data scarcity among certain participants are observed, because of imbalanced train size or data availability. Federated learning framework with synthetic data generation could solve that issue for participants with lesser data resources. 

Through experiments, it has been observed that quality of synthetic data generation is dependent on the degree of noise being added for differential privacy. For differential privacy, noise is added to model parameters as opposed to adding to original data. With federated learning, the results of experiments have shown that aggregated models after federated learning are capable of generating synthetic data of all the labels, without accessing any data for training. This further strengthens the claim of using FedSyn for non-IID data.   

\section{Future Work}
FedSyn framework is built on federated learning and GANs, with differential privacy in model parameters. Various techniques of machine learning for synthetic data generation\cite{eno2008generating, drechsler2011empirical, krishnan2016generating, jaipuria2020deflating, frid2018synthetic}, it's compatibility with federated learning could be worth exploring. 
To extend the differential privacy guarantees, the study can be extended to other techniques\cite{dwork2008differential, dwork2014algorithmic} for privacy based computing. Study can also be extended to analyze the impact of various differential privacy techniques on machine learning and deep learning, it's training efficiency, computation overhead and overall model accuracy\cite{friedman2010data, abadi2016deep}.

The research presented in this paper can also be extended further to quantify energy consumption of federated learning models, compared to centralized models, which can be the direction towards sustainable machine learning. Study can be extended to identify classes of problems, where federated learning techniques can be employed and can perform better, along with low carbon footprint. This can be further studied to identify key parameters of federated learning for energy consumption like energy consumed during communication, upload/download of model parameters, training at edge and aggregations\cite{jones2018stop, strubell2019energy, henderson2020towards, anthony2020carbontracker, qiu2021first}. Studying complex synthetic data generation methods further can even help in minimizing the energy requirements for humongous big data engineering systems across data centers\cite{jin2015significance}.

Implementation of federated learning framework with blockchain network is an active area of research, which can be extended with current study. 
This would create decentralized, trusted and secure FedSyn framework\cite{lu2019blockchain} with autonomous governance.

\section*{Disclaimer}
This paper was prepared for information purposes by the Onyx Engineering of JPMorgan Chase \& Co and its affiliates (J.P. Morgan), and is not a product of the Research Department of J.P. Morgan. J.P. Morgan makes no explicit or implied representation and warranty and accepts no liability, for the completeness, accuracy or reliability of information, or the legal, compliance, financial, tax or accounting effects of matters contained herein. This document is not intended as investment research or investment advice, or a recommendation, offer or solicitation for the purchase or sale of any security, financial instrument, financial product or service, or to be used in any way for evaluating the merits of participating in any transaction.

\section*{Acknowledgment}
The authors would like to thank the following people for their invaluable contributions and support to this project: Tulasi D Movva, Vinay Somashekhar, Thomas Eapen, Senthil Nathan, Sean Moran, Fran Silavong, Kirk Stirling and Paula Valentine from JPMorgan Chase \& Co;

\ifCLASSOPTIONcaptionsoff
  \newpage
\fi

\bibliographystyle{IEEEtran}
\bibliography{main}

\vfill

\end{document}